\documentclass[aps,twocolumn,10pt,longbibliography]{revtex4-1}

\usepackage{amssymb}
\usepackage{latexsym}

\usepackage{comment}
\usepackage{amssymb}
\usepackage{graphicx}
\usepackage{amsmath}
\usepackage{mathptmx} 
\usepackage{epstopdf}
\usepackage{color}
\usepackage{soul} 

\usepackage{url}
\usepackage{xcolor}
\definecolor{newcolor}{rgb}{.8,.349,.1}

\DeclareMathOperator\erfc{Erfc}

\begin{document}




\title{Binary matrices of optimal autocorrelations as alignment marks}


\author{Scott A. Skirlo} 
\email{sskirlo@mit.edu}
\author{Ling Lu}
\email{linglu@mit.edu}
\author{Marin Solja\v{c}i\'{c}}
\affiliation{Department of Physics, Massachusetts Institute of Technology, Cambridge, Massachusetts 02139, USA}



\begin{abstract}
We define a new class of binary matrices by maximizing the peak-sidelobe distances in the aperiodic autocorrelations.
These matrices can be used as robust position marks for in-plane spatial alignment.
The optimal square matrices of dimensions up to 7 by 7 and optimal diagonally-symmetric matrices of 8 by 8 and 9 by 9 were found by exhaustive searches.
\end{abstract}


\maketitle
\def\clr{black}

\section{Introduction}
Binary sequences~\cite{Barker1953,Neuman1971} and matrices with good autocorrelation properties have key applications in digital communications (radar, sonar, CDMA and cryptography)~\cite{Golomb2004} and in coded aperture imaging~\cite{Gottesman1989}. Several works have conducted exhaustive searches for the optimal matrices of these applications ~\cite{Alquaddoomi1989,Costas1984,Golomb1982rectangle,Mow2005}.
A less developed application of binary matrices with good aperiodic autocorrelations is two-dimensional (2D) translational spatial alignment.
For example, it has been shown in electron-beam lithography~\cite{Boegli1990} that position marks based on such binary matrices are immune to noise and manufacturing errors. However, the symbols for these applications have not been optimized~\cite{Anderson2004,Boegli1990,ling2010}. 
In this paper, we define and report the optimal binary matrices as alignment marks.
Section \ref{sec:preliminaries} sets up the problem.
Section \ref{sec:criteria} defines the criteria for the optimal matrices.
Section \ref{sec:relatedwork} discusses previous work related to this problem.
Section \ref{sec:bound} works out the useful bounds.
Section \ref{sec:search} explains the exhaustive computer searches and lists the results.
Section \ref{sec:observations} discusses several key observations of the optimal marks.
Section \ref{sec:acc} compares the performance of optimal and non-optimal marks through simulations. 
Section \ref{sec:applications} discusses the potential applications of the matrices found.
Section \ref{sec:conclusions} concludes the paper.

\section{Preliminaries}
\label{sec:preliminaries}
An alignment mark is made by creating a surface pattern different from the background so that the pattern information transforms into a two-level signal when a digital image is taken. This image can be represented as a binary matrix where 1 represents the (black) pattern pixels and 0 represents the (white) background pixels or vice versa.

The 2D aperiodic autocorrelation (A) of an $M$ by $N$ binary matrix with elements $R_{i,j}$ is defined as
\begin{equation}
A(\tau_{1},\tau_{2})=\sum\limits_{i=1}^{M}\sum\limits_{j=1}^{N}R_{i,j}R_{i+\tau_{1},j+\tau_{2}}
\end{equation}
where $\tau_{1},\tau_{2}$ are integer shifts. 
The peak value is $A(0,0)$ while all other values are sidelobes.
$A$ is an inversion-symmetric [$A(\tau_{1},\tau_{2})=A(-\tau_{1},-\tau_{2})$] $(2M-1)$ by $(2N-1)$ matrix.
The crosscorrelation between $R$ and the data image matrix $D_{i,j}$ is expressed as
\begin{equation}
C(\tau_{1},\tau_{2})=\sum_{i=1}^{M}\sum_{j=1}^{N}R_{i,j}D_{i+\tau_{1},j+\tau_{2}} .\\
\end{equation}
When the data $D$ is a noisy version of the reference $R$, the peak value of the crosscorrelation determines the most probable  position of the mark.

It is important to note that all the matrices are implicitly padded with 0s for all the matrix elements of indices exceeding their matrix dimensions. 

A linear transformation of the data matrix results in a linear transformation of the correlation as long as the reference matrix is kept the same. This can be seen from
\begin{align}
D_{i,j}^{\prime} &=cD_{i,j}+d\\
C^{\prime}(\tau_{1},\tau_{2}) &= cC(\tau_{1},\tau_{2})+d\sum^{M}_{i=1}\sum^{N}_{j=1}R_{i,j}
\end{align}
where the second term of $C^{\prime}$ is a constant. The data matrix can thus be arbitrarily scaled ($c\neq 0$) while keeping the correlation equivalent and the alignment results identical.

\textcolor{\clr}{
\section{Criteria for the optimal binary matrices}
\label{sec:criteria}
Depending on the quantities being optimized, the criteria for the optimal matrices are different.  
For alignment purposes, we list two criteria here. The first is to minimize the misalignment probability. The second is to minimize the misalignment deviation. 
The first criteria depends on the values of the autocorrelation sidelobes, while the second criteria also depends on their positions relative to the central peak.
}

\textcolor{\clr}{
In this paper, we chose to minimize the probability that misalignment happens.
A misalignment occurs when one of the sidelobes exceeds the central peak~[$p=A(0,0)$]; this probability is analytically expressed in \ref{App:math}.
Under the same noise condition, the less the peak-sidelobe distance the higher the misalignment probability. Consequently the criteria for ranking the matrices is based on their peak-sidelobe distances.
}

\begin{figure}
\centering
		\includegraphics[angle=0, width=7cm]{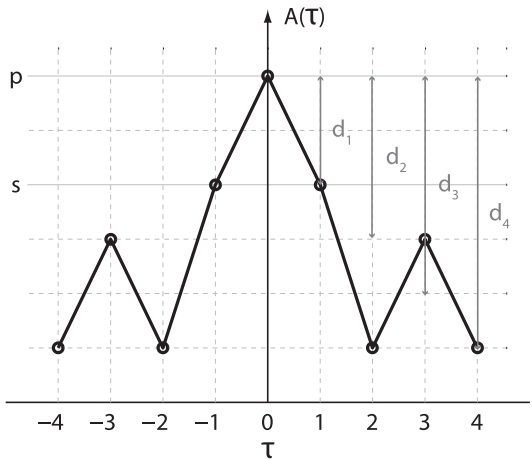}
	\caption{We illustrate an autocorrelation function $A(\tau)$, whose peak value is $p$, highest sidelobe value is $s$, and whose peak-sidelobe distances are  $d_{i}$.}
	\label{fig:illustrate}
\end{figure}

\textcolor{\clr}{
The histogram of an autocorrelation matrix can be expressed as $\{d_1|{n_1,n_2,...,n_{(s+1)}}\}$.
Here we denote the shortest peak-sidelobe distance as $d_{1}$, where $d_{1}=p-s$ and $s$ is the highest sidelobe value. The other distances are defined as $d_{i+1}=d_{i}+1$ for $i\geq{}1$. $n_{i}$ gives the number of times $d_{i}$ occurs in the autocorrelation and $\sum\limits_{i}n_i=(2M-1)(2N-1)-1$. We illustrated these definitions in Fig. \ref{fig:illustrate}.
}

\textcolor{\clr}{
The criteria for finding the optimal matrix is to maximize $d_1$ then minimize $n_i$ sequentially in the dictionary order.
This criteria is completely justified in the low noise limit in \ref{App:math} , although a general criteria depends on the amount of noise in the data matrices.
\emph{Matrices of any size} can be compared using this criteria.
In general, the distances~($d_i$) of the autocorrelation increase with the size of the matrix. Without restricting the matrix dimension, the optimal matrix will diverge in size. Consequently, we study the optimal matrix for each \emph{fixed dimension}.
Interestingly, the optimal matrices found in this paper are unique as discussed in Sec. \ref{sec:observations}.
}

\textcolor{\clr}{
\section{Related Work}
\label{sec:relatedwork}
Previous works on 1 and -1 matrices with 0 background~\cite{Alquaddoomi1989,Mow2005} in digital communications are different than our work on 1 and 0 matrices. The former representation has three levels (1,-1,0) while our binary matrices have only two levels. The aperiodic autocorrelations of these matrices are not equivalent.}


Other works on binary matrices of 1s and 0s with aperiodic autocorrelations have used different criterias selected for applications in radar and sonar. In the Costas-array problem~\cite{Costas1984}, only one black pixel is placed per column and row and the maximum sidelobe is fixed to one. In the Golomb-Rectangle problem~\cite{Golomb1982rectangle}, the number of black pixels is maximized with the restriction that the sidelobe still be fixed to one~\cite{Robinson1997}. However, our criteria does bear some resemblance to those in some of the works on one dimensional -1 and 1 (three levels) sequences~\cite{Neuman1971}. 

\section{Two Upper bounds of $d_{1,\textnormal{max}}(p)$, $d_{1,\textnormal{max}}^{\textnormal{upper,I}}(p)$ and $d_{1,\textnormal{max}}^{\textnormal{upper,II}}(p)$}
\label{sec:bound}
\begin{figure}[b]
\centering
		\includegraphics[angle=0, width=7cm]{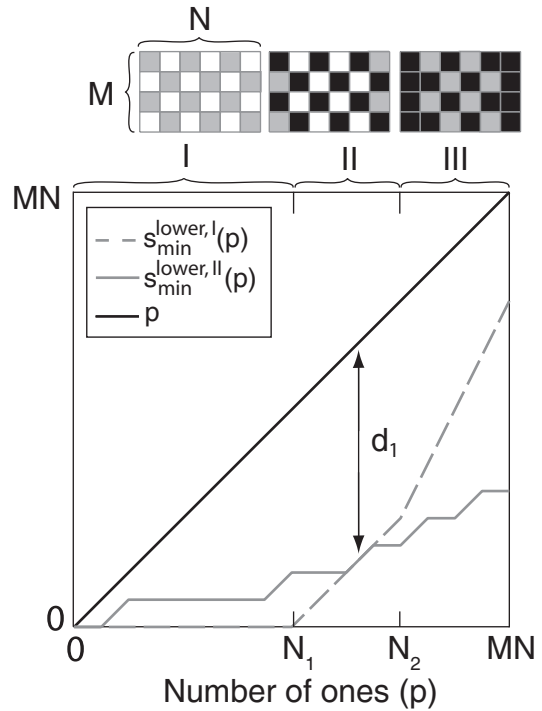}
	\caption{Lowerbounds of $s_{\textnormal{min}}(p)$, $s_{\textnormal{min}}^{\textnormal{lower,I}}(p)$ and $s_{\textnormal{min}}^{\textnormal{lower,II}}(p)$. $p$ is the autocorrelation peak. The three matrices on top illustrate the methods of filling black pixels for regions I, II and III for the matrix construction of $s_{\textnormal{min}}^{\textnormal{lower,I}}(p)$. The grey pixels show spots to be filled in that region, while the black pixels are spots that have been filled in previous regions. }
	\label{fig:bound}
\end{figure}
For a binary matrix $R$, the peak value $p$ of its autocorrelation $A$ equals the number of ones in the matrix ($R$). The largest $d_{1}$ for all matrices with a given $p$, of a fixed dimension, is $d_{1,\textnormal{max}}(p)$. $d_{1,\textnormal{max}}(p)=p-s_{\textnormal{min}}(p)$, where $s_{\textnormal{min}}(p)$ is the minimum highest sidelobe value as a function of $p$. 

In this section, we constructed an upperbound of $d_{1,\textnormal{max}}(p)$, $d_{1,\textnormal{max}}^{\textnormal{upper,I}}(p)$, by maximizing $p-A(\pm 1,0)$. The $A(\pm 1,0)$ computed here forms a lower bound on $s_{\textnormal{min}}(p)$, $s_{\textnormal{min}}^{\textnormal{lower,I}}(p)$. This construction is illustrated in Fig. \ref{fig:bound}, where we assume the matrix $R$ used to construct our bound is of dimension $M\times N$ with $M\le{}N$. 

We find:
\begin{equation}
d_{1,\textnormal{max}}^{\textnormal{upper,I}}(p) = \left\{
     \begin{array}{llc}
       p, &p \in [0,N_{1}] &\textrm{I} \\
       N_{1}, &p \in [N_{1},N_{2}] &\textrm{II} \\
       M(N+1)-p, &p \in [N_{2},MN] &\textrm{III}
     \end{array}
   \right.
\end{equation}
where $N_{1}=\frac{MN}{2}, N_{2}=\frac{MN}{2}+M$ when $MN$ is even and $N_{1}=\frac{MN+1}{2}, N_{2}=\frac{MN+1}{2}+M-1$ when $MN$ is odd.

This upperbound can be derived by starting out with a matrix $R_{i,j}=0$ for all $(i,j)$ and `filling in' with ones in a particular pattern. In region I, ones can be placed anywhere in $R_{i,j}$ where $i+j$ is odd. When $p=N_{1}$, we have formed a ``checkerboard pattern''. In region II, we place ones wherever $i+j$ is even for $i=1$ or $i=N$. In region III, the remaining locations without ones are filled.  

The autocorrelation function $A(\tau_1,\tau_2)$ equals the number of black squares that are connected by a displacement vector $(\tau_1,\tau_2)$. We can use this property to construct a second lower bound $s_{\textnormal{min}}^{\textnormal{lower,II}}(p)$. This approach is similiar to the method used in Ref.~\cite{Robinson1997}.

Since the autocorrelation is invariant under inversion, there are $((2M-1)(2N-1)-1)/2=2NM-N-M$ unique non-zero displacements; a matrix of $p$ ones fills $p(p-1)/2$ of them. As $p$ increases, there are repeated displacements because $p(p-1)/2$ quickly exceeds $2NM-N-M$. 

We can find a lowerbound $s_{\textnormal{min}}^{\textnormal{lower,II}}(p)$ by assuming that the displacements added to the autocorrelation function distribute uniformly, that is $|A(\tau_{1},\tau_{2})-A(\tau'_{2},\tau'_{2})|\le{}1$ for nonzero displacements. This gives $s_{\textnormal{min}}^{\textnormal{lower,II}}(p)=ceil[\frac{p(p-1)}{4NM-2N-2M}]$, where $ceil[x]$ rounds to the nearest integer greater than $x$. 
Consequently,  $d_{1,\textnormal{max}}^{\textnormal{upper,II}}=p-ceil[\frac{p(p-1)}{4NM-2N-2M)}]$. 

As illustrated in Fig. \ref{fig:bound}, $s_{\textnormal{min}}^{\textnormal{lower,II}}(p)$ is a better bound for small p, while $s_{\textnormal{min}}^{\textnormal{lower,I}}(p)$ is a better bound for large p.


\section{Exhaustive computer searches for the optimal square matrices}
\label{sec:search}
\begin{figure*}
   \centering
		\includegraphics[angle=0, width=15cm]{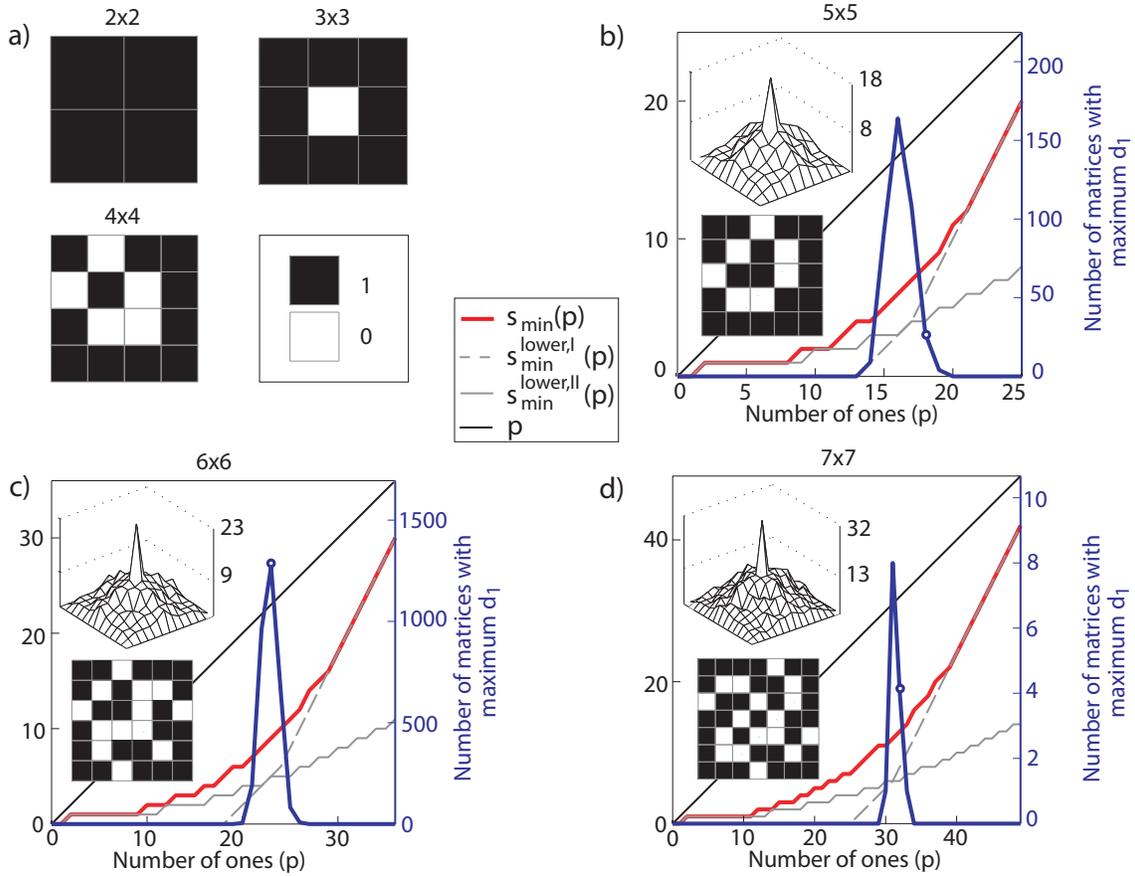}
\caption{ Results of the exhaustive searches for 2 by 2 to 7 by 7 matrices. a) The optimal matrices from 2 by 2 to 4 by 4 are shown.
b), c) and d) $s_{\textnormal{min}}(p)$ is plotted in red. The solid grey line is $s_{\textnormal{min}}^{\textnormal{lower,I}}(p)$ while the dotted grey line is $s_{\textnormal{min}}^{\textnormal{lower,II}}(p)$.
The number of the matrices having the maximum $d_{1}$ are plotted in blue.
The circle specifies the location of the optimal matrix. The optimal matrices are presented as insets below their autocorrelations, which are labeled with their $p$ and $s$ values.}
	\label{fig:2-7}
\end{figure*}

\begin{center}
\begin{figure*}
  \centering
		\includegraphics[angle=0,width=17cm]{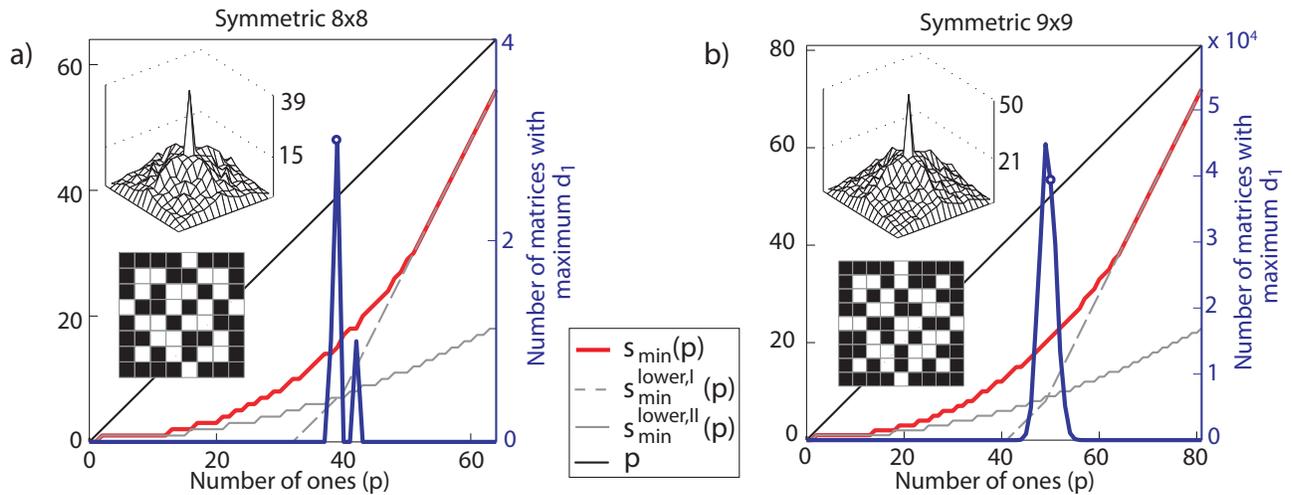}
	\caption{Results of the exhaustive searches for diagonally-symmetric 8 by 8 and 9 by 9 matrices. }
	\label{fig:symm89}
\end{figure*}
\end{center}

Physical in-plane alignment usually requires equal alignment accuracies in both directions; this calls for square matrices ($M=N$). We applied exhaustive searches to find the square matrices with the maximum $d_{1}$[$=max(d_{1,\textnormal{max}}(p))$] . The resulting matrices were ranked using the criteria in Sec. \ref{sec:criteria} to obtain the optimal matrices.

Backtrack conditions based on symmetries and sidelobes have been found useful in exhaustive searches for binary matrices ~\cite{Alquaddoomi1989,Robinson1997,Shearer2004}. Matrices related by symmetry operations are considered the same matrix. The symmetry operations for square matrices are horizontal and vertical flips and rotations by multiples of 90 degrees. For this study, a backtrack condition based on eliminating redundant matrices related by horizontal flips was implemented. Backtrack conditions based on sidelobe levels are useful if the sidelobes are being minimized. However, we are maximizing the peak-sidelobe distance $d_{1}$, so the sidelobe backtrack condition was not used. 

The search algorithm we implemented works by exhaustively generating matrices row by row. The algorithm continues generating rows until a backtrack condition occurs, or a matrix is completely specified. The matrix is stored for later ranking if it has the same or greater $d_{1}$ than the existing maximum $d_{1}$.

Several techniques were implemented to speed up the algorithm. Each matrix row was represented as a binary word so that fast bit-wise operations could be used. In addition lookup tables were created to calculate the horizontal flips and correlations of rows. For our binary matrices, the maximum sidelobes were typically located near the autocorrelation peak. Because of this, the sidelobe values were checked in a spiral pattern around the peak to quickly determine if a matrix had a $d_{1}$ less than the stored maximum.  
  
The search results for square matrices of size up to 7 by 7 are presented in Fig. \ref{fig:2-7}. Fig. \ref{fig:2-7}a) gives the optimal matrices for 2 by 2, 3 by 3 and 4 by 4.
In Fig. \ref{fig:2-7}b), c) and d) we plot, in red, $s_{\textnormal{min}}(p)$ for matrices of sizes 5 by 5, 6 by 6 and 7 by 7. This red curve is indeed bounded from below by the grey $s_{\textnormal{min}}^{\textnormal{lower,I}}(p)$ and $s_{\textnormal{min}}^{\textnormal{lower,II}}(p)$ constructed in Sec. \ref{sec:bound}.
The number of the matrices having the maximum $d_{1}$ is plotted in blue. This curve peaks around the intersection of the $d_{1,\textnormal{max}}^{\textnormal{upper,I}}$ and $d_{1,\textnormal{max}}^{\textnormal{upper,II}}$ upperbounds.
The circle on the blue line specifies the location of the optimal matrix ranked first by the criteria in Sec. \ref{sec:criteria}.
The optimal matrices and their autocorrelations are shown as insets. The two numbers on the y-axes of the autocorrelation plots are the $p$ and $s$ values of the optimal matrices. The matrices ranked second and third and their distance spectra are listed in \ref{App:spectra}.

The runtime for 7 by 7 matrices was 3 hours on 1000 Intel EM64T Nodes with 2.6 GHz clock speed. Exhaustive searches of square matrices of size 8 by 8 are not accessible to us, since the size of the search space increases exponentially with the number of matrix elements as $2^{N^2}$.

\section{Observations on the optimal square matrices}
\label{sec:observations}
The first interesting observation is that most top-ranked matrices in Fig. \ref{fig:2-7} and \ref{App:spectra} are \emph{diagonally symmetric}. 
Because of this if we restrict our searches to symmetric matrices of larger sizes, we still expect to find top-ranked matrices ~\cite{Shearer2004}. The search results for diagonally-symmetric matrices of 8 by 8 and 9 by 9 are presented in Fig. \ref{fig:symm89}. 

The second observation for our optimal matrices shown in Fig. \ref{fig:2-7}, is that $d_{1}$ always occurred in the first four neighbors of the autocorrelation peak~[$A(0,\pm{}1), A(\pm{}1,0)$]. \textcolor{\clr}{Since $d_1$ is the most likely point for misalignment,  these matrices, although optimized for misalignment probability, also have low misalignment deviation discussed in Sec. \ref{sec:criteria}. Another interesting property of autocorrelation is that the ratio of $\frac{A(0,0)-A(\pm{}1,0)}{N}$ or $\frac{A(0,0)-A(0,\pm{}1)}{N}$ is invariant under symbol expansion (i.e. expanding the number of pixels making up the original marker pixel). This property allows us to define a new quantity for the optimal matrices in this work called \emph{sharpness} $\Lambda=\frac{d_{1}}{N}$.
Since $\Lambda$ is scale-invariant, $d_1$ can be easily obtained for different scaling factors and used to evaluate the alignment performance.
The sharpness~($\Lambda$) of the optimal matrices increases with the size of the matrices.}

The third observation is that all of the optimal matrices shown in Figs. \ref{fig:2-7} and \ref{fig:symm89} are connected through their black pixel (1s) and all but 3 by 3 are connected through their white pixels. A pixel is connected if one or more of its eight neighboring pixels has the same value. \emph{Connectedness} is a preferred topological property for alignment marks; it makes the marks self-supportive, suspendible and robust against disturbances.

\textcolor{\clr}{
The fourth observation is that the optimal matrices found in Figs. \ref{fig:2-7} and \ref{fig:symm89} are \emph{unique}; there is only one matrix with the optimal histogram.
In general, the mapping from histograms to correlations is not unique. For example the 2 by 2 matrices of 
$\left[ {\begin{array}{cc}   1 & 1 \\   0 & 0 \\  \end{array} } \right]$ and
$\left[ {\begin{array}{cc}   1 & 0\\   0 & 1 \\  \end{array} } \right]$ have identical histograms.
It is unclear whether this property holds for optimal matrices of all sizes.
}
\section{Alignment accuracies of the optimal matrices}
\label{sec:acc}

\begin{figure}
\centering
		\includegraphics[angle=0, width=7cm]{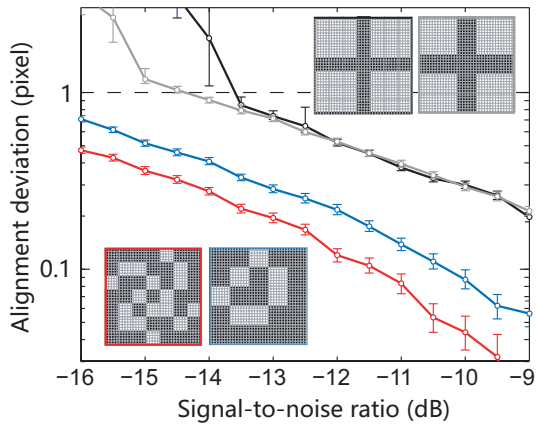}
	\caption{The ``horizontal" alignment deviation is shown for the four alignment marks under various signal-to-noise ratios. The vertical deviation is almost identical. The color of each plot line borders the corresponding marker. All markers have been expanded to 35 by 35 pixels to illustrate the idea of pixel expansion. The top, black line, on the right edge, corresponds to the 7 by 7 cross, while the second to top, grey line corresponds to the 5 by 5 cross. The second to bottom, blue line corresponds to the optimal 5 by 5 marker, while the bottom, red line corresponds to the optimal 7 by 7 matrix. }
	\label{fig:alignment}
\end{figure}

We study the performance of the optimal matrices by comparing the optimal alignment marks to the cross patterns. The matrices were embedded in a white ``0" background with a size 5 times that of the symbol. Uniform Gaussian noise was added to all pixels to simulate a noisey image.  This was correlated with its noise-free version. The alignment accuracy was determined by the deviation of the correlation peak from the center for 10000 trials.

In Fig. \ref{fig:alignment}, we plot the alignment deviation as a function of signal-to-noise ratio for two optimal marks from Fig. \ref{fig:2-7} and the crosses. The y-axis is the horizontal alignment deviation in pixels while the x-axis is the signal-to-noise ratio in decibels ($=20log{\frac{S}{N}}$). At a signal-to-noise ratio of 0 dB, the markers are barely discernible by eye. All markers were expanded to the same area, of 70 by 70 total pixels, for direct comparison. 

Applying the criteria from Section \ref{sec:criteria}, using the expanded 70 by 70 symbols, the 7 by 7 mark is ranked first, followed by the 5 by 5 mark, and then the crosses. The quality of the optimal alignment marks should improve with increasing size, which provides a motivation to continue the search for larger optimal matrices.

\section{Applications}
\label{sec:applications}
Correlation detection from a digital image is a simple, efficient and reliable way to determine the position of an alignment mark. In practice, the crosscorrelations can be calculated by fast-Fourier-transforms. The peak of the correlation can further be interpolated to obtain an alignment accuracy better than the distance represented by a single pixel of the image~\cite{Anderson2004}. The matrices reported in this paper are the desirable patterns to use in this context; they can replace the cross-type patterns widely in use today as position markers. Alignment using these matrices is very robust against noise in the imaging system and partial damage of the mark, providing the strongest peak signal for accurate sub-pixel interpolation. The potential applications of the matrices found in this paper include, but are not limited to, electron-beam lithography~\cite{Boegli1990}, planar alignment in manufacturing~\cite{Sakou1989}, synchronization~\cite{Scholtz1980} and digital watermarking~\cite{Tirkel1998}.

\textcolor{\clr}{
\section{Conclusions}
\label{sec:conclusions}
We introduced a new class of binary matrices (two level signals) which have maximial peak-to-sidelobe distances in their aperiodic autocorrelation. Optimal square matrices of dimensions up to 7 by 7 and optimal diagonally-symmetric matrices of 8 by 8 and 9 by 9 were found using a backtrack algorithm. Useful bounds, notable properties and the performances of the optimal matrices were discussed.
}
\section{Acknowledgments}
We would like to thank John D. O'Brien, Robert A. Scholtz, Yuan Shen, Moe Win, Ramesh Raskar and Steven G. Johnson for useful discussions.
This work used the Extreme Science and Engineering Discovery Environment (XSEDE), which is supported by National Science Foundation grant number OCI-1053575.
S.S. was supported by the MIT Undergraduate Research Opportunities program (UROP).
This work was supported in part by the U.S.A.R.O. through the ISN, under Contract No. W911NF-07-D-0004.
L.L. was supported in part by the MRSEC program of the NSF under Award No. DMR-0819762. 
L.L. and M.S. were partially supported by the MIT S3TEC Energy Research Frontier Center of the Department of Energy under Grant No. DE-SC0001299.



\appendix
\color{\clr}{
\section{Probability of misalignment}
\label{App:math}
The crosscorrelation between the data image and the reference matrix is denoted as $C(\tau_{1},\tau_{2})$. The autocorrelation of the binary reference matrix is denoted as $A(\tau_{1},\tau_{2})$.
The data image is essentially a copy of the reference matrix with noise added to it. We assume the noise is Gaussian and the standard deviation for each pixel is $\sigma$. The ``black" and ``white" pixel values of the data image are denoted as $b_i$ and $w_i$, whose expectation values are $\overline{b_{i}}=1$, $\overline{w_{i}}=0$ and $\overline{C}=A$.\\

Misalignment happens if $C(0,0)-C{(\tau_{1},\tau_{2})}=x_{\tau_1,\tau_2}\leq0$, representing a sidelobe~[$C{(\tau_{1},\tau_{2})}$] exceeding the central peak~[$(C(0,0))$] in the crosscorrelation. Below we write this inequality in detail,
\begin{eqnarray}
\label{eqn:inequality}
x_{\tau_1,\tau_2}=C(0,0)-C{(\tau_{1},\tau_{2})}=&  \nonumber\\
\sum_{i=1}^{p}b_{i} - \left\{ \sum_{i=1}^{p-d_{\tau_1,\tau_2}}b^{(\tau_{1},\tau_{2})}_i + \sum_{i=1}^{d_{\tau_1,\tau_2}}w^{(\tau_{1},\tau_{2})}_i\right\}  \leq 0\\
p=A(0,0), d_{\tau_1,\tau_2}=A(0,0)-A(\tau_{1},\tau_{2}) > 0 \nonumber
\end{eqnarray}

The first term in the inequality represents $C(0,0)$, where each element of the reference matrix with value 1 multiplies the corresponding $b_i$. The sum includes all $p$ pixels of $b_i$.
The two terms in the brackets represent $C{(\tau_{1},\tau_{2})}$, when the reference and data matrices are offset by ($\tau_1,\tau_2$). 
$b^{(\tau_{1},\tau_{2})}_i $ is a subset of $b_i$ which multiply elements of value 1 in the reference matrix.
$w^{(\tau_{1},\tau_{2})}_i $ is a subset of $w_i$ which multiply the remaining elements of value 1 in the reference matrix.\\

$x_{\tau_{1},\tau_{2}}$ is a sum of Gaussian variables and so is also a Gaussian variable with an expectation value $\overline{x_{\tau_1,\tau_2}}=d_{\tau_{1},\tau_{2}}$. By bookkeeping the terms in Eq. \ref{eqn:inequality}, one finds the standard deviation $\sigma_{x_{\tau_1,\tau_2}}^2=2d_{\tau_1,\tau_2}\sigma^2$.\\

The probability of misalignment due to the sidelobe at $(\tau_{1},\tau_{2})$ is $M(x_{\tau_1,\tau_2}\leq0)$.
\begin{eqnarray}
M({x_{\tau_1,\tau_2}\leq0} \mid \overline{x_{\tau_1,\tau_2}}=d_{\tau_{1},\tau_{2}})=\nonumber\\
\int_{-\infty}^{0}\frac{1}{\sqrt{2\pi}\sigma_{x_{\tau_1,\tau_2}} }\exp[\frac{-(x_{\tau_1,\tau_2}-d_{\tau_{1},\tau_{2}})^{2}}{2\sigma_{x_{\tau_1,\tau_2}}^2}]dx_{\tau_1,\tau_2}\nonumber\\
=\frac{1}{2}\erfc ( {\frac{\sqrt{d_{\tau_1,\tau_2}}}{2\sigma}})=M(\frac{d_{\tau_1,\tau_2}}{\sigma^2})\nonumber
\end{eqnarray}
Here, the complemantary error function is $\erfc(t)=\frac{2}{\sqrt{\pi}}\int_{t}^{\infty}dt'\exp{(-t'^{2})}$.

The probability of misalignment ($PoM$) is the union of the probability in the spaces bounded by all the inequalities~($x_{\tau_{1},\tau_{2}}\leq0$) at sidelobe positions $(\tau_1,\tau_2 \neq 0,0)$.
The individual spaces bounded by the inequalities overlap in general making the exact calculation of $PoM$ difficult. However, it is easy to find an upper bound for the $PoM$ by assuming no overlap between these spaces. Specifically, $PoM \leq \sum_{(\tau_{1},\tau_{2}\neq{}0,0)}{M(x_{\tau_1,\tau_2}\leq0)}$~\cite{Neuman1971} where the sum is over all sidelobes.\\

$M(\frac{d_{\tau_1,\tau_2}}{\sigma^2})$ decrease as the distance $d_{\tau_1,\tau_2}$ increases. Consequently a good criteria should tend to maximize the overall $d_i$ in order to minimize the probability of misalignment.
Also, it is of higher priority to maximize the smaller distance, which contributes more to the $PoM$.
This is the basis of our ranking criteria, which is completely justified in the low noise limit.
Under the low noise limit, the terms of larger $d_i$ make vanishingly small contributions compared to the term of smaller $d_i$. We show this in Eq. \ref{eqn:vanishing} by noticing that  $\erfc(t)$ can be approximated by $\frac{2}{\sqrt{\pi}}\frac{\exp{(-t^2)}}{t}$ for large $t$~(or small $\sigma$).
%
\begin{eqnarray}
\label{eqn:vanishing}
\lim_{\sigma \to 0} \frac{M(d_{i+1}/\sigma^2)}{M(d_{i}/\sigma^2)}
=\lim_{\sigma \to 0} \exp{[-\frac{d_{i+1}-d_{i}}{2\sigma^2}]}\frac{\sqrt{d_{i}}}{\sqrt{d_{i+1}}}=0
\end{eqnarray}
However, the ranking criteria, in general, depends on the noise level $\sigma$.
We note, due to the central limit theorem, the above results still hold for non-Gaussian noise distributions, when the matrix size is large.
}
\clearpage
\color{black}{}
\section{Distance spectra}
\label{App:spectra}
\begin{figure}[h]
\centering
		\includegraphics[width=0.3\textwidth]{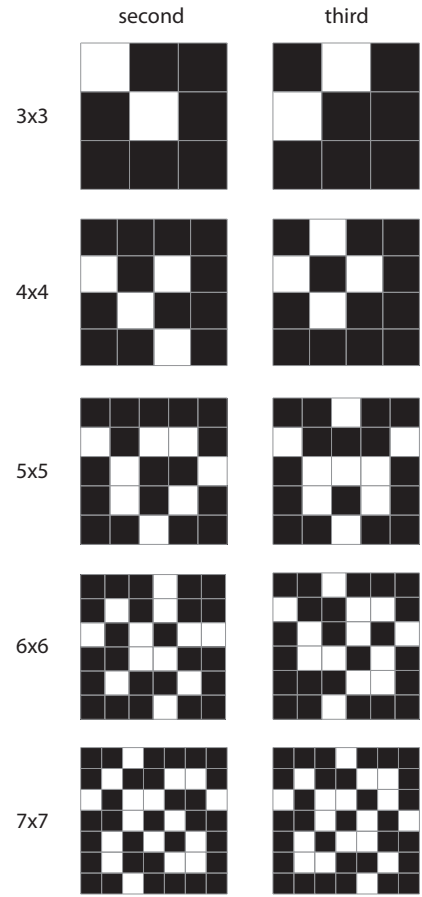}
	\caption{Matrices ranked second and third. The first-ranked optimal matrices are shown in Fig. \ref{fig:2-7}.}
	\label{fig:2nd3rdMatrices}
\end{figure}

In order to provide additional useful matrices and to illustrate our ranking criteria, we tabulated, in Table \ref{tab:spectra},  part of the peak-sidelobe distance spectra for the top-three ranked square matrices from the exhaustive search results. The values of the first four distances ($d_1,d_2,d_3,d_4$) and the numbers ($n_1,n_2,n_3,n_4$) of the corresponding sidelobes are listed.
Those top-three binary square matrices are shown in Fig. \ref{fig:2-7} and in Fig. \ref{fig:2nd3rdMatrices}.

\begin{table}[h]
	\centering
		\caption{Peak-sidelobe distance spectra of the top-three ranked square matrices from the exhaustive search results.}
	\label{tab:spectra}
		\begin{tabular}{|l|cccc|}
			\hline
        $\textrm{N}\times\textrm{N} $ & $d_1$ & $d_2$ & $d_3$ & $d_4$\\
			\hline
        $\textrm{Ranking}$ & $n_1$ & $n_2$ & $n_3$ & $n_4$\\
		  \hline\hline
        $3\times 3$ & $ 4 $ & $ 5 $ & $ 6 $ & $ 7 $\\
        \hline
         First   & 4    &  4   & 12  & 4  \\
         Second   & 4    &  12   & 6  & 2  \\
         Third   & 6    &  6   & 12  & 0  \\
			\hline\hline
        $4\times 4$ & $ 7 $ & $ 8 $ & $ 9 $ & $ 10 $\\
        \hline
         First   & 8    &  8   & 22  & 10  \\
         Second   & 10    &  2   & 10  & 18  \\
         Third   & 12    &  0   & 8  & 20  \\
			  \hline\hline
        $5\times 5$ & $ 10 $ & $ 11 $ & $ 12 $ & $ 13 $\\
        \hline
         First   & 4    &  6   & 10  & 6  \\
         Second   & 4    & 12   & 8  & 6  \\
         Third   & 4    &  12   & 16  & 12  \\
			  \hline\hline
        $6\times 6$ & $ 14 $ & $ 15 $ & $ 16 $ & $ 17 $\\
        \hline
         First   & 4   &  16   & 4  & 2  \\
         Second   & 6    &  6   & 12  & 4  \\
         Third   & 6    &  8   & 12  & 6  \\
			  \hline\hline
        $7\times 7$ & $ 19 $ & $ 20 $ & $ 21 $ & $ 22 $\\
        \hline
         First   & 14    &  8   & 6  & 0  \\
         Second   & 16    &  4   & 4  & 4  \\
         Third   & 16    &  4   & 8  & 4  \\
			  \hline
		\end{tabular}
	
\end{table} 
\newpage






\end{document}